\documentclass[letterpaper]{article} 
\usepackage{aaai2026}  
\usepackage{times}  
\usepackage{helvet}  
\usepackage{courier}  
\usepackage[hyphens]{url}  
\usepackage{graphicx} 
\usepackage{natbib}  
\usepackage{caption} 
\usepackage[ruled,linesnumbered]{algorithm2e}
\usepackage{algorithmicx}
\usepackage{amssymb}
\usepackage{multirow}

\usepackage{graphicx}
\usepackage{amsmath}
\usepackage{tcolorbox}
\usepackage{enumitem}

\usepackage{xcolor}

\usepackage{newfloat}
\usepackage{listings}
\DeclareCaptionStyle{ruled}{labelfont=normalfont,labelsep=colon,strut=off} 
\title{CATCH: A Controllable Theme Detection Framework  \\ with Contextualized Clustering and Hierarchical Generation}
\author{
    Rui Ke~\textsuperscript{\rm 1}, Jiahui Xu~\textsuperscript{\rm 1}, Shenghao Yang~\textsuperscript{\rm 1}, Kuang Wang~\textsuperscript{\rm 1}, Feng Jiang~\textsuperscript{\rm 2}\thanks{Corresponding Author.}, Haizhou Li~\textsuperscript{\rm 1,3,4} \\
}
\affiliations{
    \textsuperscript{\rm 1}Shenzhen Research Institute of Big Data, The School of Data Science, The Chinese University of Hong Kong, Shenzhen\\
    \textsuperscript{\rm 2}Artificial Intelligence Research Institute, Shenzhen University of Advanced Technology\\
    \textsuperscript{\rm 3}The School of Artificial Intelligence, The Chinese University of Hong Kong, Shenzhen\\
    \textsuperscript{\rm 4}Department of Electrical and Computer Engineering, National University of Singapore\\
    jiangfeng@suat-sz.edu.cn
}

\usepackage{bibentry}

\begin{document}

\maketitle

\begin{abstract}
Theme detection is a fundamental task in user-centric dialogue systems, aiming to identify the latent topic of each utterance without relying on predefined schemas. Unlike intent induction, which operates within fixed label spaces, theme detection requires cross-dialogue consistency and alignment with personalized user preferences, posing significant challenges. Existing methods often struggle with sparse, short utterances for accurate topic representation and fail to capture user-level thematic preferences across dialogues. To address these challenges, we propose CATCH (Controllable Theme Detection with Contextualized Clustering and Hierarchical Generation), a unified framework that integrates three core components: (1) context-aware topic representation, which enriches utterance-level semantics using surrounding topic segments; (2) preference-guided topic clustering, which jointly models semantic proximity and personalized feedback to align themes across dialogue; and (3) a hierarchical theme generation mechanism designed to suppress noise and produce robust, coherent topic labels. Experiments on a multi-domain customer dialogue benchmark (DSTC-12) demonstrate the effectiveness of CATCH with 8B LLM in both theme clustering and topic generation quality. 
\end{abstract}

\section{Introduction}

\begin{figure}[!t]
\centering
\includegraphics[width=\linewidth]{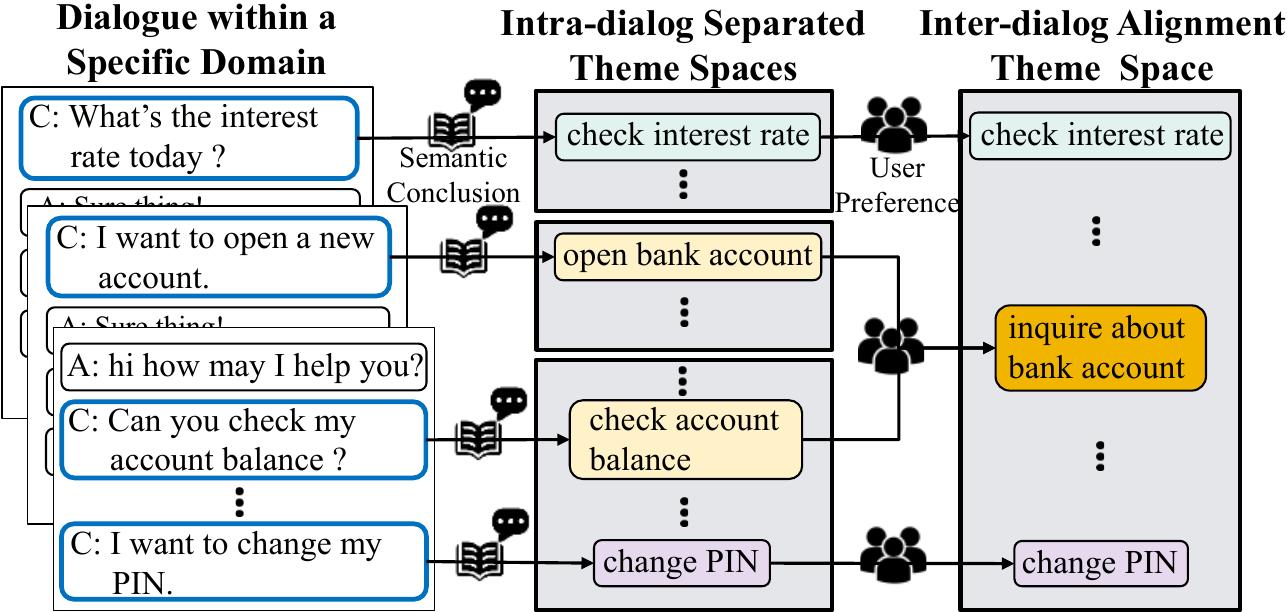}
\caption{Illustration of the controllable theme detection task. Given a set of dialogues with unlabeled utterances, a theme is generated for each utterance. The theme granularity is influenced by auxiliary inputs such as user preferences ~\cite{dstc12}.}
\label{fig:task_demo}
\end{figure}

In real-world customer service domains such as banking, finance, travel, and insurance, accurately identifying the underlying theme of each user utterance is essential for enhancing service efficiency, understanding user needs, and retrieving contextually relevant knowledge. Unlike intent induction~\cite{gung2023intent}, which typically maps utterances to a predefined label space~\cite{pu2022dialog, costa2023towards}, theme detection aims to uncover latent and potentially novel topics without prior knowledge. Effective theme detection requires preliminary precise topic assignment within a single dialogue~\cite{nguyen2022adaptive, du2013topic}, but more importantly should be consistent across multiple dialogues and align with user preferences~\cite{dstc12}, which regularizes inter-dialogue theme consolidation, as illustrated in Figure~\ref{fig:task_demo}. These challenges underscore the need for models that can generalize beyond surface-level semantics and adapt to diverse real-world conversational scenarios.

However, existing approaches fail to address the real-world controllable theme detection for three key challenges. First, short utterances often lead to sparse and ambiguous semantic signals, making it difficult for conventional topic modeling methods~\cite{blei2003latent, pham2024neuromax} to construct reliable topic representations. Second, while topic clustering methods~\cite{chatterjee2020intent, gung2023intent} group utterances based on surface-level semantics, they typically overlook user-specific preferences, resulting in inconsistent clustering across dialogues even when the underlying intent is similar. Moreover, most previous work lacks a structured and controllable theme generation mechanism~\cite{perkins2019dialog, zeng2021automatic}, causing the generated topic labels to vary arbitrarily between contexts and limiting their applicability in downstream applications.

To address these challenges, we propose \textbf{CATCH}~(\textbf{C}ontrollable \textbf{A}nd \textbf{T}hematic \textbf{C}lustering with \textbf{H}ierarchy), a controllable theme detection framework that combines intra-dialogue context modeling with inter-dialogue user preference alignment. Specifically, CATCH comprises three key components: (1) a context-aware topic representation module that enhances semantic expressiveness by leveraging dialogue-level topic segmentation; (2) a preference-guided clustering mechanism that jointly models semantic similarity and user-specific signals to ensure thematic consistency across dialogues; and (3) a hierarchical theme generation process, inspired by Chain-of-Thought prompting and stabilized via majority voting, that yields robust, structured, and domain-adaptive topic labels.

Experimental results on the Controllable Dialogue Theme Detection benchmark (DSTC-12) show that our CATCH consistently outperforms strong baselines in both in-domain and cross-domain settings, even under a few preference supervisions. Notably, our framework ranks \textbf{second} in the official blind evaluation, achieving strong performance in both automatic and human assessments with a lightweight design (about an 8B-level model). Extensive ablation studies and qualitative analyses further confirm the effectiveness and generalizability of each component. The main contributions of this work are as follows:

\begin{itemize}
\item We propose \textbf{CATCH}, a novel controllable theme detection framework that integrates topic segmentation, preference-enhanced clustering, and hierarchical label generation. It enables fine-grained and user-adaptive theme modeling across dialogues, bridging structured representation with user-centered controllability.
  
\item We introduce a contextualized clustering mechanism that jointly models intra-dialogue semantic flow and inter-dialogue user preference patterns, thereby aligning topic grouping not only with content coherence but also with personalized thematic expectations.

\item We propose a hierarchical theme generation strategy guided by controllable LLM prompting, where local semantic candidates are first derived within fine-grained clusters, and subsequently consolidated into coherent themes through consensus-driven aggregation, mitigating label drift and enhancing cross-context consistency.

\item Our framework achieves \textbf{second place} in the DSTC-12 Controllable Theme Detection track under both automatic and human evaluations, with a lightweight design and high sample efficiency. Extensive ablation studies and qualitative analyses confirm its effectiveness, particularly in low-resource and cross-domain settings.
\end{itemize}

\section{Related Work}

\subsection{Intra-dialogue Theme Detection}
Intra-dialogue theme detection focuses on identifying the thematic affiliation of each utterance within a single conversation. It generally involves two subtasks: \emph{topic segmentation} and \emph{topic generation}.

\textbf{Topic segmentation}.
Dialogue Topic Segmentation aims to divide a dialogue into semantically coherent segments. Early unsupervised approaches relied on surface-level signals such as word co-occurrence~\cite{hearst-1997-text, eisenstein-barzilay-2008-bayesian} or topic distributions~\cite{riedl-biemann-2012-topictiling, du-etal-2013-topic}. More recent methods construct contrastive utterance-pair data and fine-tune pretrained models like BERT~\cite{vuong2025hicot, devlin2019bert, xing2021improving, gao2023unsupervised} and LLMs~\cite{fan2024uncovering, xu2024unsupervised}. However, these methods apply uniform segmentation strategies across datasets with different granularities, often leading to inconsistent performance due to domain mismatch and dialogue fragmentation.

\textbf{Topic generation}.
Topic generation methods, often based on probabilistic or neural topic modeling~\cite{blei2003latent, pham2024neuromax}, aim to infer high-level abstract themes from dialogue utterances. Yet, the brevity and sparsity of dialogue turns pose challenges for robust topic inference~\cite{bach2021dynamic, lin2024combating}. To address this, several methods attempt to augment short texts via auxiliary signals~\cite{nguyen2022adaptive, tuan2020bag, jiang2024advancing}. Despite their success in extracting global themes, most existing methods fail to ensure intra-dialogue consistency or control the granularity of the generated labels.

\subsection{Inter-dialogue Theme Detection}

Inter-dialogue theme detection mainly concerns clustering and aligning topics across multiple dialogues, which is critical to understanding user intent on scale and constructing reusable dialogue themes.

A common strategy is to cluster utterances based on their semantic representations~\cite{nguyen2025glocom, grootendorst2022bertopic, zhang2022neural, sia2020tired}. While these methods are efficient and scalable, they often assume a fixed set of topics and do not support theme discovery~\cite{perkins2019dialog}. Other works explore the complexity of theme space via multiview clustering~\cite{vu2025topic, perkins2019dialog}, density-based methods~\cite{chatterjee2020intent}, or structured prediction signals~\cite{liu2021open, zeng2021automatic}. However, most of these approaches focus solely on semantic similarity and overlook the influence of user preferences on thematic variation. This leads to inconsistent topic grouping across dialogues, even when the underlying intent is similar.

Furthermore, none of the above approaches provides explicit or controllable mechanisms for theme label generation, which limits their applicability in settings that require interpretable and stable outputs across domains.

\begin{figure*}[!t]
\centering
\includegraphics[width=\textwidth]{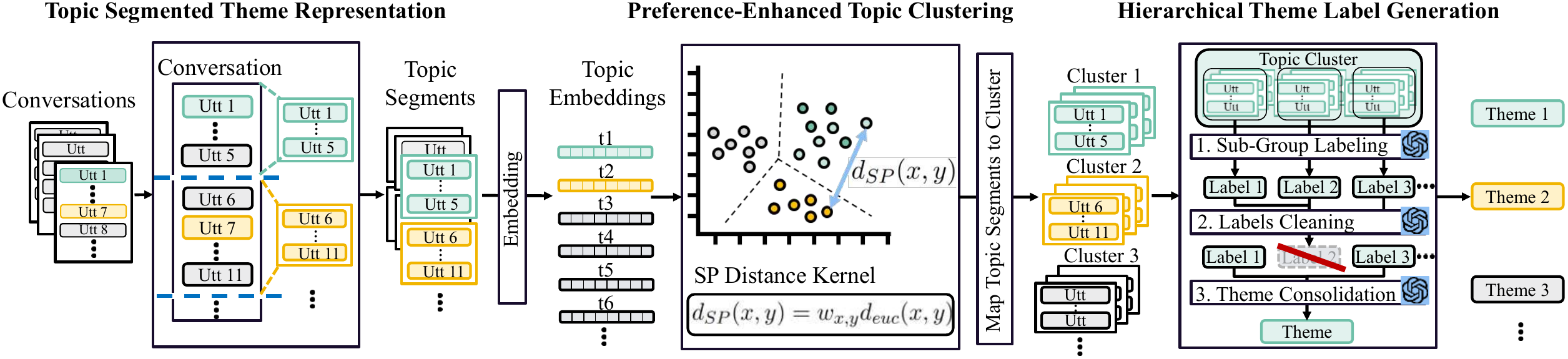}
\caption{The overall workflow of CATCH.}
\label{fig:work_flow}
\end{figure*}

\section{Methodology}

\subsection{Task Definition}
We define the controllable theme detection (TD) task as a structured theme generation problem over dialogue utterances. Given a set of utterances $U = \{u_1, \dots, u_m\}$ extracted from dialogues of a specific domain with low resource user preference annotation (as shown in Table \ref{tab:dataset}), the goal is to assign each utterance $u_i \in U$ a theme label $L_i$ that is both preference-aligned and contextually consistent across dialogues. Formally, the output of the theme detection process is a finite family of theme labels $\mathbb{L} = \{L_1, \dots, L_n\}$ aligned with user preferences $P$, where each label corresponds to a subset of utterances:
\begin{equation}
\{L_1:\{u_j, \dots, u_k\}, \dots, L_n\} \gets TD(U \mid P)
\end{equation}

To achieve this goal, as illustrated in Figure~\ref{fig:work_flow}, we propose \textbf{CATCH}, a controllable theme detection framework that incorporates both intra- and inter-dialogue modeling~\footnote{https://github.com/SUAT-AIRI/CATCH}.

\subsection{Context-Aware Intra-Dialogue Theme Representation} \label{sec:top_seg}

To address the semantic sparsity and ambiguity commonly observed in short utterances, we design a context-aware intra-dialogue theme representation module. It leverages a dual-branch topic segmentation framework to infer latent segment boundaries and construct thematically coherent spans. Each utterance representation is then enriched by aggregating contextual information from its corresponding segment, enabling more robust and consistent theme encoding across dialogue turns.

\paragraph{Dual-branch Topic Segmentation Framework.} Inspired by DialSTART~\cite{gao2023unsupervised}, our topic segmentation module adopts a dual-branch structure composed of two parallel encoders: (1) a topic encoder to measure topical similarity between utterances in a dialogue and (2) a coherence encoder to model discourse continuity between utterance-pairs. These two components collaborate to detect topic boundaries in a multi-turn dialogue. Specifically, for each adjacent utterance-pair $(u_i, u_{i+1})$, the segmenter computes a boundary score $r_i$ by aggregating the outputs of both encoders:
\begin{equation}
B = \{ r_i = t_i + c_i \mid i = 1, \dots, n-1 \}
\end{equation}\label{eql:B} Here, $t_i$ is the cosine similarity between the topic representations $\mathbf{h}_i$ and $\mathbf{h}_{i+1}$ from the Topic Encoder, while $c_i$ is the coherence score generated by the Coherence Encoder.

\paragraph{Contrastive Objectives for Encoder-Specific Learning.} To optimize the segmenter, we formulate two contrastive losses based on the outputs of the two encoders. These losses enable the model to distinguish thematically or coherently similar utterances from unrelated ones.

(1) Topic Encoder Contrastive Loss $L_1$: This loss supervises the Topic Encoder to promote topical consistency in a dialogue. For each utterance $u_i$, positive samples $S_i$ are drawn from a window of neighboring utterances, and negative samples ${S}_i^\star$ are sampled from outside this window:
\begin{equation}
L_1(u_i) = \sum_{u^+ \in S_i} \sum_{u^- \in S^*_i} \max(0, \eta + e_i^- - e_i^+)
\end{equation}
where $e_i^+$ and $e_i^-$ represent the similarity scores between $u_i$ and the positive ($u^+$) and negative ($u^-$) samples, respectively, and $\eta$ is a margin hyperparameter.

(2) Dual Encoder Contrastive Loss $L_2$: This loss supervises both the Coherence and Topic Encoder to capture discourse-level continuity in different dialogues. For each utterance-pair $P_i=(u_i,u_{i+1})$, we define positive utterance-pair segments $\mathbb{P}_i$ (neighboring segment pairs) and negative pairs $\mathbb{P}_i^\star$ (randomly sampled from the other dialogues). The loss is defined as:
\begin{equation}
L_2(P_i) = \sum_{P^+ \in \mathbb{P}_i} \sum_{ P^-\in \mathbb{P}^*_i} \max(0, \eta + r_i^- - r_i^+)
\end{equation} with the segment-level relevance score $r_i^\pm$ computed by $P_i=(u_i,u_{i+1})$ and $P^{\pm} = (u,v)^\pm$:
\begin{equation}
r_i = \text{sim}\left(\frac{h_{i}+h_{i+1}}{2}, \frac{h^u+{h}^v}{2}\right) + c_i
\end{equation} where $h$ is the topic representations of utterance provided by Topic Encoder, and the $\pm$ of $r_i$ is the same with $\pm$ of $P$.

The final optimization combines both encoder-specific contrastive learning objectives:
\begin{equation}
\mathcal{L} = \frac{1}{N} \sum_{i=1}^{N} L_1(u_i) + \frac{1}{M} \sum_{i=1}^{M} L_2(P_i)
\end{equation}
where $N$ and $M$ denote the number of samples for each loss.

\paragraph{Two-Stage Adaptation Strategy.} We further enhance the model's ability to generalize and align with the target domain through a two-stage adaptation process:

\textbf{Stage 1: Conversation-Level Adaptation.} This stage aims to adapt the model to the target dialogue domain by constructing training samples from every utterance $u_i$ in the target dataset. For each $u_i$, both the Topic Encoder and Coherence Encoder are supervised using the aforementioned contrastive objectives $L_1$ and $L_2$.

\textbf{Stage 2: Utterance-Level Adaptation.} To further specialize the model for real-world usage scenarios, we conduct fine-grained adaptation based on utterances annotated with themes. We assume these annotated utterances are treated as likely indicators of topic boundaries and closer to the real-world using scenarios.

After adaptation, we apply the TextTiling algorithm~\cite{hearst-1997-text} on the computed boundary scores $B$ to segment the dialogue. Given a target utterance, we extract its surrounding segment as its contextual block and use it to construct a semantically grounded theme representation $T$, which serves as the input to the following topic clustering.

\begin{figure}[!t]
\centering
\includegraphics[width=\linewidth]{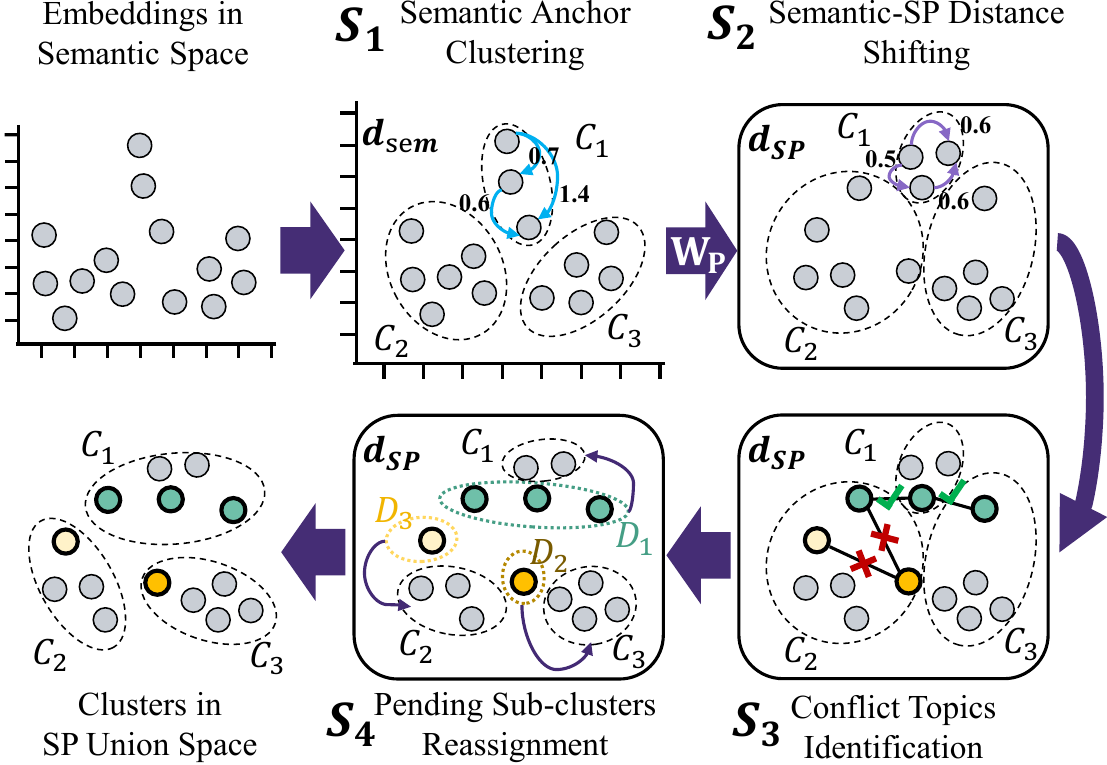}
\caption{Illustration of the Preference-Enhanced Topic Clustering Algorithm. ``$\color{teal}\checkmark$'' denotes a strong linking preference, while ``$\color{red}\times$'' denotes a strong splitting preference.}
\label{fig:cluster_alg}
\end{figure}

\subsection{Preference-Enhanced Topic Clustering}

To acquire the theme family $\mathbb{L}$, we aim to group similar topics across different dialogues into coherent theme clusters leveraging on their theme representations $T$ obtained by the topic segmentor. To this end, we propose a preference-enhanced topic clustering algorithm that jointly considers semantic similarity and user preference signals, guided by a trained preference reward model (PRM).

We first design a Semantic-Preference (SP) distance kernel to dynamically fuse semantic similarity with user preference. For each topic pair $(x, y)$, the SP is defined as:
\begin{equation}
    d_{SP}(x, y) = w_{x,y} \cdot d_{sem}(x, y)
\end{equation}
where $d_{sem}$ denotes the Euclidean distance between topic embeddings, and $w_{x,y} = PRM(x, y)$ is a preference scalar produced by the reward model to indicate the degree to which two topics belong to the same theme.

Because the true joint space combining semantic and preference information is latent and not explicitly constructed, we propose a four-step algorithm grounded in semantic space but progressively incorporating preference signals, as illustrated in Figure~\ref{fig:cluster_alg}.

\begin{itemize}
    \item \textbf{Step 1 (S1):} Topics are clustered solely based on semantic similarity to form initial anchor clusters, such as $C_I$.
    \item \textbf{Step 2 (S2):} The mutual SP distance $d_{SP}$ of topics are calculated and assigned to substitute semantic distance $d_{sem}$ by multiplying Preference scalars, such as $W_p$ delegating the full matrix of $w_{x,y}$.
    \item \textbf{Step 3 (S3):} Preference scalars are further used to detect \emph{conflict} topic pairs whose preference signals contradict their current cluster assignments. These conflict topics are then extracted and re-clustered in the SP distance space to form new pending sub-clusters, such as $D_I$.
    \item \textbf{Step 4 (S4):} Each pending sub-cluster is reassigned to the most compatible anchor cluster by minimizing the aggregated SP distance, thereby producing final preference-consistent clusters.
\end{itemize}

Furthermore, since annotating user preferences across all topic pairs is costly and inefficient, we train a preference reward model to estimate human preferences between utterances. The model is trained on a small number of preference-aligned samples ($S$), each labeled as either belonging to the same theme or not, and is used to assign compatibility scores to topic pairs.

\begin{equation}
S = \{ ((t_i, t_j), y_{ij}) \mid y_{ij} \in \{0,1\} \}
\end{equation}
We regard the Preference Reward Model (PRM) as a classification model that predicts a continuous preference score $w_{ij} = \text{PRM}(t_i, t_j) \in [0,1]$ indicating the degree of linking tendency between the pair. The model is optimized using mean squared error (MSE) loss:
\begin{equation}
\mathcal{L}_{\text{PRM}} = (w_{ij} - y_{ij})^2
\end{equation}

\begin{algorithm}
\caption{Hierarchical Theme Generation}\label{alg:labeling_1}
\KwData{K topic clusters $C_K$, where each $C_i \in C_K$ containing topic segment $P_j$ as elements; Prompt-based LLM $LLM(prompt|input)$}
\KwResult{Themes of topic clusters $\{T_{K}\}$}

\For{each $C_i$}{
    Randomly split $C_i$ into several groups $S_N$ with $S_n = \{P^n_1,...,P^n_{10}\}$ and $\uplus_{n \in N} S_n = C_i$ \;
    \For{each $S_n$}{
        $l_n \gets LLM(Label|S_n)$;
    }
    Initialize the cleaned labels set $L_i$ \;
    $l^{core}  \gets LLM(VoteCore|\{l_n\},n \in N)$ \;
    \For{each $l_n$}{
        \If{$LLM(IsRelevant|(l_n,l^{core}))$}{
            $L_i.append(l_n)$ \;
        }
    }
    $T_i \gets LLM(Conclude|L_i)$ \;
}
\end{algorithm}

\subsection{Hierarchical Theme Label Generation}
In this final step, each topic cluster $C_i$ is assigned a structured label $T_i$. To support downstream applications requiring domain-adaptive and interpretable labeling, we employ a prompt-based large language model (LLM) to generate concise, actionable theme names in the form of verb phrases that reflect user intent.

As illustrated in Algorithm~\ref{alg:labeling_1}, the three-stage hierarchical generation framework, integrating fine-grained local labeling with semantic consolidation, is an autonomous data negotiation process allowing automatic theme emergence.

\textbf{Intra-group Labeling.}
Following a divide-and-conquer strategy, each topic cluster is partitioned into smaller groups to satisfy LLM input length constraints and enhance generation quality. The diversity across groups also helps retain semantic variation within the cluster. Then, each group is labeled through the LLM prompt ($IsRelevant$).

\textbf{Noisy Label Cleaning.}
Group-level labels may include inconsistent or noisy descriptions. To filter these, we prompt LLM to identify a core label $l^{core}$ via majority voting ($VoteCore$) which selects the most frequent group label, and retain only those labels deemed semantically relevant ($IsRelevant$) to $l^{core}$, resulting in a cleaned label set $L_i$.

\textbf{Inter-group Theme Consolidation.}
Finally, the cleaned label set $L_i$ is summarized using a consolidation prompt ($Conclude$) to yield the final theme label $T_i$, which captures the dominant intent of the cluster while maintaining semantic consistency across utterances.
 
\section{Experiments}

\subsection{Experimental Settings} \label{sec:settings}

\textbf{Datasets}. We conduct experiments on the multi-domain customer support dialogue datasets (Banking, Finance, Insurance, and Travel) provided by DSTC-12~\cite{dstc12}, as summarized in Table~\ref{tab:dataset}. Each dataset contains two key types of annotations of the themed utterances:
(1) Theme Annotation: Each target utterance is annotated with its corresponding theme label. (2) Preference Annotation: A binary relation (\textit{should-link} and \textit{cannot-link}) of a pair of target utterances indicating whether they should be grouped under the same theme (should-link) or not (cannot-link).

In the offline evaluation, we use the Banking dataset for training and the Finance and Insurance domains as the valid dataset. For online evaluation, we deploy our model to predict theme labels on the Travel dataset, which lacks golden annotations. The predicted results are submitted to the organizers of DSTC-12 for blind evaluation. Throughout training, CATCH is trained solely based on preference annotations without accessing the ground-truth theme labels. 

\begin{table}[!ht]
    \centering
    \resizebox{\linewidth}{!}{
    \begin{tabular}{lllll}
        \hline
        \textbf{Type} & \textbf{Domain} & \textbf{\# Dialog.} & \textbf{\# Utterance} & \textbf{\# Preference}\\
        \hline
        Offline & Banking & 1634 & 58418 (980) & 164/164 \\
        Offline & Finance & 1725 & 196764 (3000) & 173/173 \\
        Offline & Insurance & 836 & 60352 (1333) & 155/126 \\
        Online  & Travel & 765 & 72010 (999) & 77/77\\
        \hline
    \end{tabular}
    }
   \caption{Data Statistics of the DSTC-12 Dataset. The numbers in parentheses indicate the number of sampled utterances with annotated themes. In the \textit{Preference} column, the values denote the number of should-link / cannot-link utterance pairs, respectively.}
    \label{tab:dataset}
\end{table}

\begin{table*}[!ht]
    \centering
    \resizebox{\linewidth}{!}{
    \begin{tabular}{l|cc|ccc|cc|ccc}
        \hline
        \textbf{Method} 
        & \multicolumn{5}{c|}{\textbf{Finance}} 
        & \multicolumn{5}{c}{\textbf{Insurance}} \\
        \hline
        & \multicolumn{2}{c|}{\textbf{Clustering Metrics}} 
        & \multicolumn{3}{c|}{\textbf{Theme Label Quality}} 
        & \multicolumn{2}{c|}{\textbf{Clustering Metrics}} 
        & \multicolumn{3}{c}{\textbf{Theme Label Quality}} \\
        & \textbf{Acc} & \textbf{NMI} 
        & \textbf{Rouge-1/2/L} & \textbf{CosSim} & \textbf{LLM-Score} 
        & \textbf{Acc} & \textbf{NMI} 
        & \textbf{Rouge-1/2/L} & \textbf{CosSim} & \textbf{LLM-Score} \\
        \hline
        GRand 
        & 24.6 & 28.2 & 5.0 / 3.5 / 5.0 & 13.8 & 87.0 
        & 41.5 & 42.2 & 12.3 / 0.0 / 12.3 & 47.8 & 86.6 \\
        
        GRM 
        & 39.1 & 51.5 & 21.6 / 6.4 / 21.1 & 42.8 & 82.9 
        & 39.6 & 51.7 & 27.1 / 8.4 / 26.2 & \textbf{57.5} & 96.5 \\

        GSP 
        & 23.3 & 28.0 & 19.1 / 4.1 / 19.0 & 48.5 & 85.8 
        & 23.5 & 30.1 & 20.8 / 8.3 / 20.7 & 44.6 & 87.2 \\

        \hline
        \textbf{CATCH} 
        & \textbf{55.8} & \textbf{67.1} & \textbf{42.4} / \textbf{24.5} / \textbf{42.4} & \textbf{59.3} & \textbf{97.3} 
        & \textbf{54.5} & \textbf{62.6} & \textbf{41.8} / \textbf{16.1} / \textbf{41.8} & 57.0 & \textbf{100.0} \\
        \hline
    \end{tabular}
    }
    \caption{Performance comparison on Finance and Insurance domains, grouped by theme distribution and theme label quality. All values are in percentages.}
    \label{tab:out_results}
\end{table*}

\begin{table}[!ht]
    \centering
    \resizebox{\linewidth}{!}{
    \begin{tabular}{l|lllr}
        \hline
        \textbf{Stage} 
        & \textbf{Model} & \textbf{Size} 
        & \textbf{Training Cost} \\
        \hline
        TopSeg
        & 2*Bert-base & 220M 
        & 2*GPU·h \\
        \hline
        
        PeC
        & all-mpnet-base-v2 & 110M/109M 
        & 0.05/0*GPU·h  \\
        \hline

        HieGen 
        & LLaMA-3-8B & 8B 
        & 0*GPU·h \\
        
        \hline
    \end{tabular}
    }
    \caption{Computational cost (4090). TopSeg: Topic segmentation state; PeC: preference-enhanced clustering stage; HieGen: hierarchical theme generation stage.}
    \label{tab:computational_cost}
\end{table}

\textbf{Metrics}. 
To comprehensively evaluate the effectiveness of \textsc{CATCH}, we follow the DSTC-12~\cite{dstc12} evaluation protocol, which assesses two core aspects: (1) the quality of theme segmentation (i.e., utterance clustering), and (2) the quality of generated theme labels.

\textit{Offline Evaluation.}
For theme segmentation quality, we use two standard clustering metrics: 
\textbf{Normalized Mutual Information (NMI)}~\cite{NMI}, which quantifies the mutual dependence between predicted and reference clusters normalized by their entropies, and 
\textbf{Clustering Accuracy (Acc)}, computed via the Hungarian algorithm to align clusters optimally. For theme label quality, we evaluate the semantic and textual correspondence between predicted and reference labels using:
\textbf{Cosine Similarity (CosSim)} based on Sentence-BERT embeddings, 
\textbf{ROUGE}~\cite{ROUGE} for n-gram overlap, and 
an \textbf{LLM-based score} that assesses label format and informativeness via vicuna-13B evaluation guided by human-crafted criteria.

\textit{Online Evaluation.}
For the held-out test set without golden labels, the DSTC-12 organizers perform additional evaluations including both automatic metrics and manual human judgments.

\textbf{Baselines}. 
We compare \textsc{CATCH} against three representative baselines: (1) \textbf{GRand}: the official DSTC-12 baseline, which performs utterance-level clustering using randomly sampled should-link and cannot-link pairs, followed by direct theme generation.
(2) \textbf{GRM}: an enhanced variant of GRand that conducts topic-level clustering and uses a preference reward model to propagate preference annotations to all topic pairs before generating labels. (3) \textbf{GSP}: a variant of GRM that employs the SP distance kernel for clustering without adopting the two-step algorithm.

\subsection{Implementation Details}~\label{implementation}
In the topic segmentation stage, we follow the previous work~\cite{gao2023unsupervised}, using \textit{bert-base-uncased} and \textit{sup-simcse-bert-base-uncased} as our coherence encoder and topic encoder, respectively. During the pre-training and fine-tuning process, we both set the learning rate to 5e-6 and the epoch to 3. The relevance threshold is set to be 0.5 during boundary inference.

In the preference-enhanced topic clustering, we employ \textit{all-mpnet-base-v2} to obtain the sentence transformer embeddings and use \textbf{UMAP} to reduce embedding dimension. For semantic clustering, we employ \textbf{Spectrum} clustering method with the default cluster number K being 30, following the common design. As for PRM, we use \textit{bert-base-uncased} as the default reward model with a learning rate of 2e-5 and an epoch of 3. During preference inference, we set the tendency threshold of linking $\theta_l$ to be 0.85 and the tendency threshold of splitting $\theta_s$ to be 0.15. 

In the theme label generation, we employ \textit{LLaMA3-8B-Instruct} as the default LLM for prompt generation. The number of topics in each subgroup is set to be 25.

In the evaluation part, offline LLM-based evaluation is conducted on Vicuna-13B as DSTC-12 suggests, while the online test result is provided by DSTC-12.

The computational requirement of three stages is shown in Table \ref{tab:computational_cost}. CATCH is a lightweight design that is only trained on models with less than 500M parameters in total, and uses all substitutable open-source models.

\subsection{Offline Experimental Results}

\begin{table}[!ht]
    \centering
    \resizebox{\linewidth}{!}{
    \begin{tabular}{l|cc|ccc}
        \hline
        \textbf{Method}
         & \textbf{Acc} & \textbf{NMI} & \textbf{Rouge-1/2/L} & \textbf{Cos} & \textbf{LLM} \\
        \hline
        GRand & 36.8 & 33.4 & 11.1 / 2.9 / 11.1 & 30.8 & 82.0 \\ 
        GRM   & 46.9 & 51.6 & 22.0 / 3.8 / 20.4 & 37.3 & 86.8 \\ 
        GSP   & 15.4 & 4.4  & 6.9 / 0.6 / 6.7   & 52.8 & 90.7 \\ 
        \hline
        CATCH & \textbf{56.7} & \textbf{65.4} & \textbf{35.3} / \textbf{10.0} / \textbf{35.3} & \textbf{58.5} & \textbf{95.9} \\ 
        \hline
    \end{tabular}
    }
    \caption{Performance on the Banking Domain.}
    \label{tab:in_results}
\end{table}

We train CATCH on the Banking dataset and conduct the experiment in two data scenarios: in-domain data, out-of-domain data. In the in-domain task, we test on Banking set. For the out-of-domain task, we respectively evaluate the model on the Finance and Insurance sets.

\subsubsection{The Performance in the In-domain Scenario.}

Table~\ref{tab:in_results} demonstrates that \textsc{CATCH} achieves the best performance across all evaluation metrics, showing consistent improvements in both theme segmentation and label generation quality. For \textbf{theme distribution}, \textsc{CATCH} yields the highest Clustering Accuracy (Acc = 56.7\%, NMI = 65.4\%), significantly outperforming the strongest baseline GRM (Acc = 46.9\%, NMI = 51.6\%). This indicates that the preference-enhanced clustering provides more accurate and consistent theme segmentation. For \textbf{theme label quality}, \textsc{CATCH} also leads by a large margin with ROUGE-1 of 35.3\% and a Cosine Similarity (CosSim) of 58.5\%, surpassing GRM. Furthermore, the 95.9\% LLM-based score indicates high consistency with predefined label formats and semantics.

Interestingly, GSP obtains high CosSim (52.8\%), but with extremely low Acc (15.4\%) and NMI (4.4\%), suggesting that it generates plausible-sounding labels but poor underlying topic segmentation. This confirms that directly clustering in SP union space without a grounding space results in misleading theme segmentation.

\subsubsection{The Performance in the Out-of-domain Scenario.}
Follow DSTC-12 suggest settings, two out-of-domain datasets (Finance and Insurance) are used to evaluate the generalization ability of \textsc{CATCH} . As shown in Table~\ref{tab:out_results}, \textsc{CATCH} consistently outperforms all baselines across all metrics in both domains, demonstrating strong robustness and cross-domain adaptability. In terms of \textbf{theme distribution}, preference-enhanced clustering helps \textsc{CATCH} achieve the highest clustering scores in both Finance (Acc = 55.8, NMI = 67.1) and Insurance (Acc = 54.5, NMI = 62.6), significantly surpassing GRM and other baselines. This also indicates that our clustering mechanism remains effective even when applied to unseen domains. For \textbf{theme label quality}, \textsc{CATCH} again leads with strong ROUGE scores (e.g., ROUGE-L = 42.4 in Finance, 41.8 in Insurance) and the highest CosSim and LLM-Score due to the hierarchical theme generation paradigm. Notably, these values surpass those on the in-domain test, underscoring the strong generalization capability of our generation paradigm across domains.

In addition, contrasting all baselines, \textsc{CATCH} performs the best on the Finance domain, whose dialogues have a more ambiguous theme. We attribute this to both the topic-level representation that makes it easier to infer latent thematic structures, and the preference-enhanced clustering that introduces auxiliary input to correctly classify vague points.

\subsection{Online Blind Evaluation}

We present the official blind test results in Table~\ref{tab:auto_eval} and in Table~\ref{tab:human_eval}. Our submission (Team E), based on a lightweight LLaMA-3-8B model without any reliance on proprietary models (e.g., GPT-4o), achieved \textbf{second place} overall in both evaluations, and especially, performed stably high in theme label quality. The other two competitive works, Team C \cite{Kim2025KSTC} and Team D \cite{Lee2025limit}, both leverage proprietary LLMs more or less.

\textbf{Automatic Evaluation.} As shown in Table~\ref{tab:auto_eval}, our system ranked second overall with an average score of 67.5\%, closely following Team C. Even with a smaller model, CATCH demonstrated comparable capability to Team C in the final \textbf{label generation} task with 41.2 ROUGE-L, 62.5\% Cosine Similarity, and consistently high BERTScores (F1 = 93.3\%), though for \emph{theme distribution}, the 35.8\% Acc and 47.7\% NMI are relatively low. This indicates the robustness of CATCH in providing high-quality themes even though clustering quality is low. Notably, CATCH continuously generates structurally clean and stylistically consistent labels as shown by an average of 94.6\% in \emph{style scores}.

\begin{table*}[ht]
\centering
\resizebox{\textwidth}{!}{
\begin{tabular}{l|l|cc|ccc|ccc|c}
\hline
\textbf{Team ID} 
& \textbf{LLM}
& \textbf{Acc} & \textbf{NMI} 
& \textbf{Rouge-1/2/L} & \textbf{CosSim} & \textbf{BERTScore (P/R/F1)}  
& \textbf{Sec-1} & \textbf{Sec-2} & \textbf{Avg} 
& \textbf{Overall} \\
\hline
Team C & API 
& \textbf{68.0} & \textbf{70.4} 
& \textbf{45.2 / 23.8 / 45.1} & \textbf{69.9} & \textbf{95.0 / 94.7 / 94.7} 
& \textbf{100.0} & \textbf{99.5} & \textbf{99.7} & \textbf{75.5} \\

Team E (ours) & $<$30B 
& 35.8 & \underline{47.7} 
& \underline{42.3} / 16.5 / \underline{41.2} & \underline{62.5} & \underline{93.9 / 92.8 / 93.3} 
& \underline{93.5} & \underline{95.7} & \underline{94.6} & \underline{67.5} \\

Team D & $<$30B 
& \underline{51.8} & 47.7 
& 34.6 / \underline{21.3} / 34.3 & 55.9 & 92.5 / 91.5 / 91.9 
& 80.4 & 76.6 & 78.5 & 63.1 \\

Team A & API
& 48.4 & 42.0 
& 32.7 / 4.6 / 29.8 & 59.5 & 89.8 / 91.2 / 90.4 
& 46.0 & 56.5 & 51.2 & 53.5 \\

Team F & $<$30B 
& 26.7 & 9.1 
& 23.1 / 0.8 / 21.1 & 46.0 & 85.7 / 89.3 / 87.2 
& 4.1 & 3.5 & 3.8 & 33.4 \\

Team B & API 
& 17.9 & 2.0 
& 5.0 / 0.0 / 5.0 & 37.1 & 85.2 / 88.0 / 86.5 
& 12.0 & 0.1 & 6.1 & 28.8 \\
\hline
\end{tabular}
}
\caption{Automatic evaluation results on the blind test set (Travel domain). All values are percentages. LLM: API indicates usage of proprietary models via API; $<$30B denotes open models smaller than 30B.}
\label{tab:auto_eval}
\end{table*}

\textbf{Human Evaluation.} 
Our model also ranked second in human evaluation with an average score of \textbf{71.8\%}, being consistent with automatic evaluation, highlighting its practical effectiveness and robustness. We received high marks in both \emph{semantic relevance} (86.3\%) and \emph{domain specificity} (91.1\%), showing our labels are both meaningful and context-aware. In terms of linguistic quality, we achieved \textbf{93.7\%} in both grammatical structure and word choice, affirming our system's fluency and readability.

\begin{table*}[ht]
\centering
\resizebox{\textwidth}{!}{%
\begin{tabular}{l|ccccc|cc|c|c}
\hline
\textbf{Team ID} 
& \multicolumn{5}{c|}{\textbf{Per-Utterance Functional Metrics}} 
& \multicolumn{2}{c|}{\textbf{Per-Cluster Structural Metrics}} 
& \textbf{Per-Cluster Functional (TD)} 
& \textbf{Overall Avg.} \\
& \textbf{SR} & \textbf{AU} & \textbf{GR} & \textbf{ACT} & \textbf{DR} 
& \textbf{CWC} & \textbf{GS} 
& & \\
\hline
Team C & \textbf{89.7} & \textbf{82.8} & \textbf{47.8} & \textbf{74.8} & \textbf{98.8} & \textbf{100.0} & \textbf{100.0} & \textbf{91.1} & \textbf{85.6} \\
Team E (ours) & 86.2 & 54.6 & 22.5 & 54.5 & 91.1 & 93.7 & 93.7 & 78.3 & 71.8 \\
Team D & 68.8 & 63.7 & 26.4 & 60.3 & 94.3 & 91.7 & 66.7 & 90.9 & 70.3 \\
Team A & 77.3 & 63.7 & 22.8 & 56.2 & 79.8 & 83.3 & 100.0 & 75.8 & 69.8 \\
Team F & 45.2 & 41.6 & 7.7 & 41.6 & 67.5 & 95.0 & 100.0 & 72.6 & 58.9 \\
Team B & 65.0 & 12.9 & 0.0 & 4.1 & 97.8 & 100.0 & 33.3 & 0.0 & 39.1 \\
\hline
\end{tabular}}%
\caption{Human evaluation results on the blind test set (Travel domain). All values are percentages. Metrics: Semantic Relevance (SR), Analytical Utility (AU), Granularity (GR), Actionability (ACT), Domain Relevance (DR), Conciseness \& Word Choice (CWC), Grammatical Structure (GS), and Thematic Distinctiveness (TD).}
\label{tab:human_eval}
\end{table*}

\begin{table}[!ht]
    \centering
    \resizebox{\linewidth}{!}{
    \begin{tabular}{l|cc|ccc}
        \hline
        \textbf{Model} 
        & \textbf{Acc} & \textbf{NMI} 
        & \textbf{Rouge-1/2/L} & \textbf{Cos} & \textbf{LLM} \\
        \hline
        CATCH
        & \textbf{55.8} & \textbf{67.1} 
        & \textbf{42.4} / 24.5 / \textbf{42.4} & \textbf{59.3} & 97.3 \\
        \hline
        w/o-PeC 
        & 48.8 & 59.6 
        & 40.7 / \textbf{26.7} / 40.7 & 51.5 & \textbf{98.4} \\

        w/o-TopSeg 
        & 41.6 & 53.3 
        & 23.6 / 10.1 / 23.6 & 45.3 & 87.8 \\

        w/o-HieGen 
        & 36.8 & 48.2 
        & 19.6 / 9.1 / 19.6 & 30.3 & 82.3 \\
        \hline
    \end{tabular}
    }
    \caption{Module-Level ablation results on the Finance set.}
    \label{tab:ablation_results}
\end{table}

\subsection{Ablation Study}\label{sec:ablation_study}
To better understand the contribution of each component in our framework, we conduct comprehensive ablation studies from two perspectives: Module-level ablation and Setting-level ablation.

\subsubsection{Module-Level Ablation.} \label{sec:ablation_study_module}
We evaluate the full model and its three ablated variants on the Finance dataset (see Table~\ref{tab:ablation_results}): \textbf{w/o-PeC}: clustering only relies on semantics. \textbf{w/o-TopSeg}: theme representation is based solely on raw utterance embeddings.  \textbf{w/o-HieGen}: removes hierarchical generation and adopts flat label generation instead.

All three modules contribute substantially to overall performance. Discarding \textbf{PeC} causes dis-alignment with user preferences, shown by -7.8\% decrease in CosSim. Removing \textbf{TopSeg} significantly decreases clustering quality, with -14.2\% Acc and -13.8\% NMI, demonstrating its importance in capturing topical coherence across utterances. Simplifying \textbf{HieGen} in flat generation leads to the greatest loss in label generation quality, particularly in ROUGE-L (-2.8\%), confirming the effectiveness of hierarchical modeling.

Furthermore, w/o-HieGen causes great backward in theme distribution with -19\% Acc and -18.9\% NMI, because HieGen is always capable of assigning a correct label for the majority of topics within a mingled cluster, where flat label generation usually encounters malfunction, thus provides meaningless labels. Section: Correlation Analysis of PeC and HieGen further probes the relationship between PeC and HieGen.

\subsubsection{Setting-Level Ablation.} \label{sec:ablation_study_settings}
To evaluate the stability and design sensitivity of each module, we further examine alternative settings while keeping the overall framework intact, as shown in Table~\ref{tab:ablation_results_settings}.

We first examine the effectiveness of different training depths for the topic segmentation module by either using the original DialSTART without adaptation or only on the conversation-level adaptation dataset. It exhibits a clear trend: deeper domain adaptation leads to better performance. 

We also substitute the original settings in the other two modules with stronger ones: replace the clustering algorithm (Spectral Clustering) with ITER-DBSCAN~\cite{chatterjee2020intent}(DBS) and ClusterLLM~\cite{zhang2023clusterllm}(ClusLLM) , and LLaMA3-8B with GPT-4o for label generation. Both substitution results show the improvements in clustering and theme label quality. Moreover, the combination of both strong substitutions (ClusLLM+GPT-4o) achieves the best performance, demonstrating that our framework consistently benefits from stronger submodules, leading to comprehensive performance gains.

\begin{table}[!ht]
    \centering
    \resizebox{\linewidth}{!}{
    \begin{tabular}{l|cc|ccc}
        \hline
        \textbf{Model} 
        & \textbf{Acc} & \textbf{NMI} 
        & \textbf{Rouge-1/2/L} & \textbf{Cos} & \textbf{LLM} \\
        \hline

        CATCH
        & 55.8 & 67.1 
        & 42.4 / 24.5 / 42.4 & 59.3 & 97.3 \\
        \hline

        w/o-Utte. Ada
        & 50.6 & 59.7 
        & 35.6 / 20.0 / 35.6 & 58.7 & 95.3 \\
        
        w/o-Utte. + Conv. Ada
        & 44.3 & 56.6 
        & 32.5 / 5.9 / 29.4 & 53.0 & 90.2 \\
        \hline

        w/-DBS 
        & 60.4 & 69.2 
        & 45.7 / 27.8 / 45.7 & 61.5 & 98.1 \\

        w/-ClusLLM
        & 70.1 & 73.4 
        & 46.7 / 28.2 / 46.7 & 62.1 & 97.5 \\
        \hline

        w/-Qwen-3-8B 
        & 56.0 & 67.3 
        & 43.6 / 25.4 / 43.6 & 59.3 & 97.3 \\

        w/-GPT4o 
        & 57.0 & 68.1 
        & 45.7 / 26.2 / 44.9 & 61.1 & 98.3 \\
        \hline

        w/-DBS+GPT4o
        & 61.8 & 70.1
        & 46.8 / 28.6 / 45.9 & 63.7 & 98.6 \\

        w/-ClusLLM+GPT4o
        & \textbf{70.4} & \textbf{74.1} 
        & \textbf{49.9 / 30.3 / 48.8} & \textbf{71.0} & \textbf{98.6} \\
        \hline
    \end{tabular}
    }
    \caption{Setting-level ablation results on the Finance set.}
    \label{tab:ablation_results_settings}
\end{table}

\begin{figure}[ht]
\centering
\includegraphics[width=\linewidth]{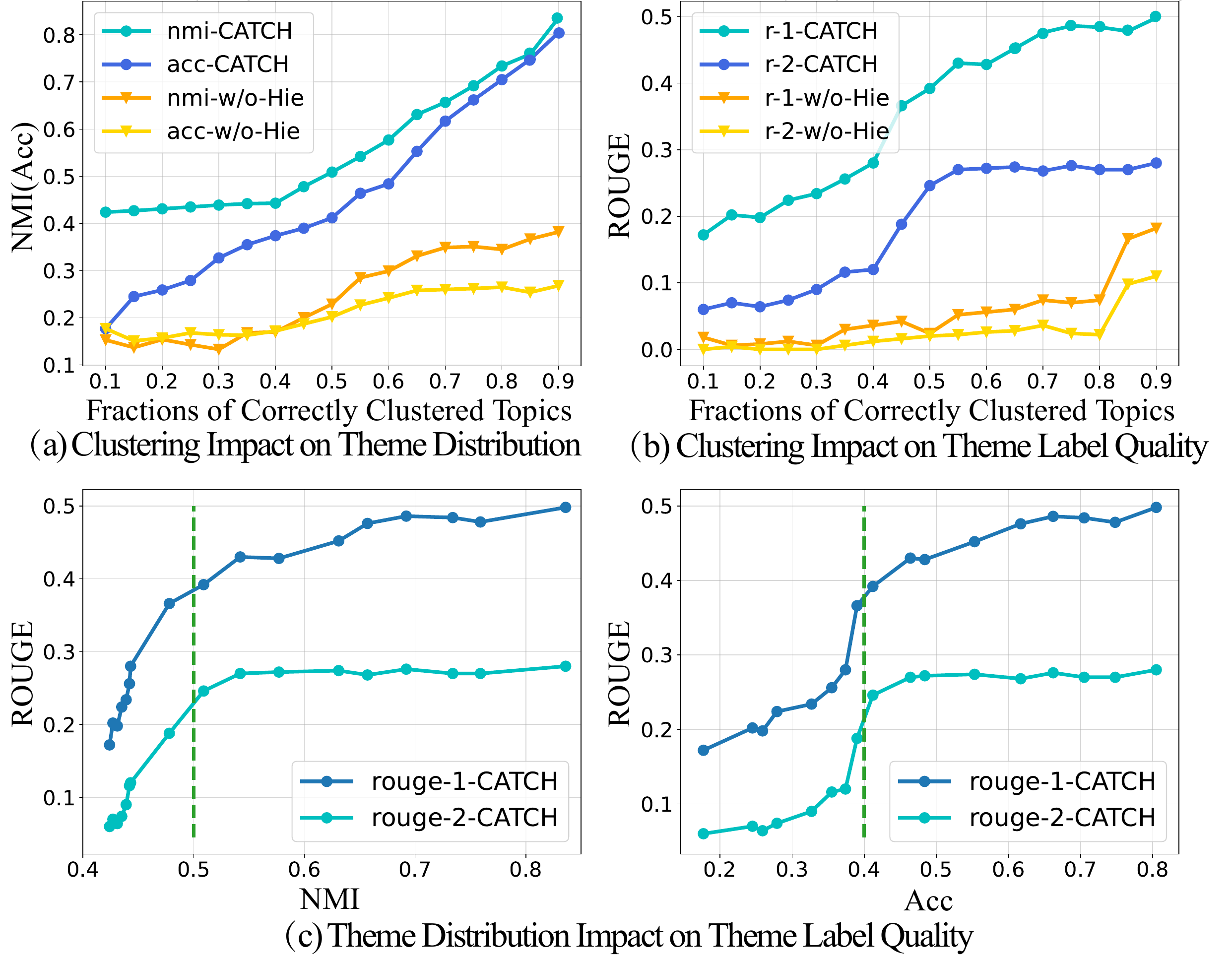}
\caption{Correlation analysis of PeC and HieGen. Figure~\ref{fig:analysis}(a) illustrates the relationship between FCC and cluster-level metrics (NMI and Accuracy).}
\label{fig:analysis}
\end{figure} 

\subsection{Analysis}
Though the ablation study suggests that HieGen can partially compensate for the instability introduced by PeC, the mutual effect between components remained unclear. Thus, we further conduct cascade error analysis to expose the error delivery process, and then analyze specifically the profound correlation between PeC and HieGen.

\subsubsection{Cascade Error Analysis.}
To decompose the aggregation error to each component, we respectively set the accuracy of TopSeg, HieGen, and their combination to be 100\%, creating three ideal models: w/-TopSeg=1, w/-PeC=1, and Both=1. Moreover, a model represents the ideal upper limit (IUL) is set to be the combination of w/-GPT4o and w/-TopSeg, PeC=1.

\begin{table}[!ht]
    \centering
    \resizebox{\linewidth}{!}{
    \begin{tabular}{l|c|cc|ccc}
        \hline
        \textbf{Model} 
        & \textbf{1-pk} & \textbf{Acc} & \textbf{NMI} 
        & \textbf{Rouge-1/2/L} & \textbf{Cos} & \textbf{LLM} \\
        \hline

        CATCH
        & 0.37 & 55.8 & 67.1 
        & 42.4 / 24.5 / 42.4 & 59.3 & 97.3 \\
        \hline

        w/-TopSeg=1
        & 1 & 56.9 & 66.3 
        & 46.9 / 26.3 / 45.3 & 66.1 & 98.4 \\

        w/-PeC=1
        & 0.37 & 100 & 100
        & 51.6 / 32.4 / 51.6 & 73.7 & 98.4 \\

        Both=1 
        & 1 & 100 & 100 
        & 56.0 / 34.5 / 55.5 & 77.9 & 98.7 \\

        IUL 
        & 1 & 100 & 100 
        & \textbf{59.0 }/ \textbf{37.4} / \textbf{58.1} & \textbf{80.2} & \textbf{99.5} \\

        \hline
    \end{tabular}
    }
    \caption{Cascade error analysis on the Finance set. 1-pk indicates the topic boundary accuracy.}
    \label{tab:cascade_error}
\end{table}
The results in Table \ref{tab:cascade_error} align with the ablation findings: reducing each partial error steadily improves final performance. Notably, the framework is more sensitive to cluster quality, yielding a larger gain (+9.2\% w/-PeC=1) compared with topic segmentation (+4.5\% w/-TopSeg=1) on Rouge-1.

\subsubsection{Correlation Analysis of PeC and HieGen.}
To further investigate the interplay between PeC and HieGen, we analyze how clustering quality affects the final theme labeling performance. Specifically, we evaluate both CATCH and its ablated variant w/o-HieGen under varying \textit{fractions of correctly clustered topics} (FCC), by randomly reassigning a portion of topics to incorrect clusters.

As shown in Figure~\ref{fig:analysis}(a) and (b), CATCH consistently outperforms w/o-HieGen in both theme distribution accuracy and label quality. HieGen exhibits strong intra-cluster denoising capabilities: the cluster-level quality remains closely proportional to FCC, while in contrast, the performance of w/o-HieGen plateaus at a low level due to its limited capacity in generating coherent theme labels.

Furthermore, Figure~\ref{fig:analysis}(b) shows that HieGen substantially reduces the system’s dependency on clustering quality. For CATCH, theme label quality improves sharply even with a large amount of clustering errors, where FCC exceeds only 0.4, while the improvement for w/o-HieGen delays to FCC surpasses 0.8. This trend is further extended in Figure~\ref{fig:analysis}(c), which demonstrates the relationship between theme distribution consistency and label quality. Notably, HieGen maintains strong performance even under suboptimal clustering: ROUGE-1 remains above 0.4 and ROUGE-2 above 0.26, despite limited NMI (0.5) and Accuracy (0.4). These results show that HieGen’s divide-and-conquer denoising can fix clustering errors as long as about half of the topics are correctly grouped.

\section{Conclusion}
In this paper, we propose CATCH, a novel theme detection framework that significantly enhances the automatic discovery and consistency of themes within a latent topic space aligned with user preferences. By treating the entire architecture as a theme generation pipeline, CATCH jointly models intra-dialogue theme representation and inter-dialogue preference-aware alignment, leading to coherent and user-aligned theme labels. Extensive experiments demonstrate the robustness and generalizability of CATCH across diverse tasks, while ablation studies further reveal the complementary roles and coordination of its three key modules. In future work, we plan to continuously improve the framework with cutting-edge techniques and make it more adaptive and dynamic, enabling its application to a broader range of downstream scenarios, such as proactive dialogue systems, dialogue control, and fine-grained dialogue analysis.

\section{Acknowledgments}
This work was supported by the project of National Natural Science Foundation of China (Grant No. 62271432), Shenzhen Science and Technology Program (Grant No. ZDSYS20230626091302006), and Program for Guangdong Introducing Innovative and Entrepreneurial Teams ( Grant No. 2023ZT10X044), Key Project of Shenzhen Higher Education Stability Support Program (Grant No. 2024SC0009), and Shenzhen Science and Technology Program (Grant No. RCBS20231211090538066).

\bibliography{aaai2026}

\appendix

\end{document}